\documentclass{article}

\usepackage{microtype}
\usepackage{graphicx}
\usepackage{subcaption}
\usepackage{booktabs} 

\usepackage{hyperref}

\usepackage[accepted]{sysml2019}

\sysmltitlerunning{Collage Inference}

\begin{document}

\twocolumn[
\sysmltitle{Collage Inference: Using Coded Redundancy for Low Variance Distributed Image Classification}




\begin{sysmlauthorlist}
\sysmlauthor{Krishna Giri Narra}{usc}
\sysmlauthor{Zhifeng Lin}{usc}
\sysmlauthor{Ganesh Ananthanarayanan}{msr}
\sysmlauthor{Salman Avestimehr}{usc}
\sysmlauthor{Murali Annavaram}{usc}
\end{sysmlauthorlist}

\sysmlaffiliation{usc}{Department of Electrical Engineering, University Of Southern California, Los Angeles, California, USA}
\sysmlaffiliation{msr}{Microsoft Research, Redmond, Washington, USA}

\sysmlcorrespondingauthor{Krishna Giri Narra}{narra@usc.edu}

\sysmlkeywords{Availability, Neural Networks, Deep Learning, Coded Computing, Redundancy, Cloud Computing, Machine Learning}
\vskip 0.3in

\begin{abstract}
MLaaS (ML-as-a-Service) offerings by cloud computing platforms are becoming increasingly popular. Hosting pre-trained machine learning models in the cloud enables elastic scalability as the demand grows.
But providing low latency and reducing the latency variance is a key requirement. Variance is harder to control in a cloud deployment due to uncertainties in resource allocations across many virtual instances.
We propose the collage inference technique which uses a novel convolutional neural network model, collage-cnn, to provide  low-cost redundancy. A collage-cnn model takes a collage image formed by combining multiple images and performs multi-image classification in one shot, albeit at slightly lower accuracy. We augment a collection of traditional single image classifier models with a single collage-cnn classifier which acts as their low-cost redundant backup. Collage-cnn provides backup classification results if any single image classification requests experience slowdown. Deploying the collage-cnn models in the cloud, we demonstrate that the 99th percentile tail latency of inference can be reduced by 1.2x to 2x compared to replication based approaches while providing high accuracy. Variation in inference latency can be reduced by 1.8x to 15x.

\end{abstract}
]




\section{Introduction}
\label{sec:introduction}
Deep learning is used across many fields such as live video analytics, autonomous driving, health care, data center management, and machine translation.  
Providing low latency and low variance inference is critical in these applications. On the deployment front machine learning as a service (MLaaS) platforms \cite{azure_ml, google_ml_engine, aws_sagemaker} are being introduced by many datacenter operators. 

Prediction serving, namely inference, on MLaaS platforms is attractive for scaling inference traffic. Inference requests can be served by deploying the trained models on the MLaaS platforms. To achieve scalability of the prediction serving systems, incoming queries are distributed across multiple replicas of the trained model. As the inference demands grow, an enterprise can simply increase the cloud instances to meet the demand. However, virtualized and distributed services are prone to slowdowns, which lead to high variability and long tail in the inference latency. Slowdowns and failures are more acute in cloud-based deployments because of the widespread sharing of compute, memory and network resources~\cite{tail}.

Existing straggler mitigation techniques can be broadly classified into three categories: replication \cite{tail, late}, approximation \cite{approx}, coded computing \cite{speedUpML, LMA_all, li2016fundamental}. In replication based techniques, additional resources are used to add redundancy during execution: either a task is replicated at it's launch or a task is replicated on detection of a straggler node. Approximation techniques ignore the results from tasks running on straggler nodes. Coded computing techniques add redundancy in a coded form at the launch of tasks and have proven useful for linear computing tasks. In deep learning several of these techniques have been studied for mitigating stragglers in training phase. However these solutions need to be revisited when using MLaaS for inference. For example, replicating every request pro-actively as a straggler mitigation strategy could lead to significant increase in resource costs. Replicating a request reactively on the detection of a straggler, on the other hand, can increase latency.  


In this work we argue that, while prediction serving using MLaaS platforms would be more prone to slowdowns, they are also more amenable to low-cost redundancy schemes. Prediction serving systems deploy a front-end load balancer that receives requests from multiple users and submits them to the back-end cloud instances. In this setting, the load balancer has the unique advantage of treating multiple requests as a single collective and create a more cost effective redundancy strategy.

\begin{figure}
    \centering
    \begin{subfigure}{0.52\textwidth}
        \centering
        \includegraphics[width=\textwidth, height=4cm]{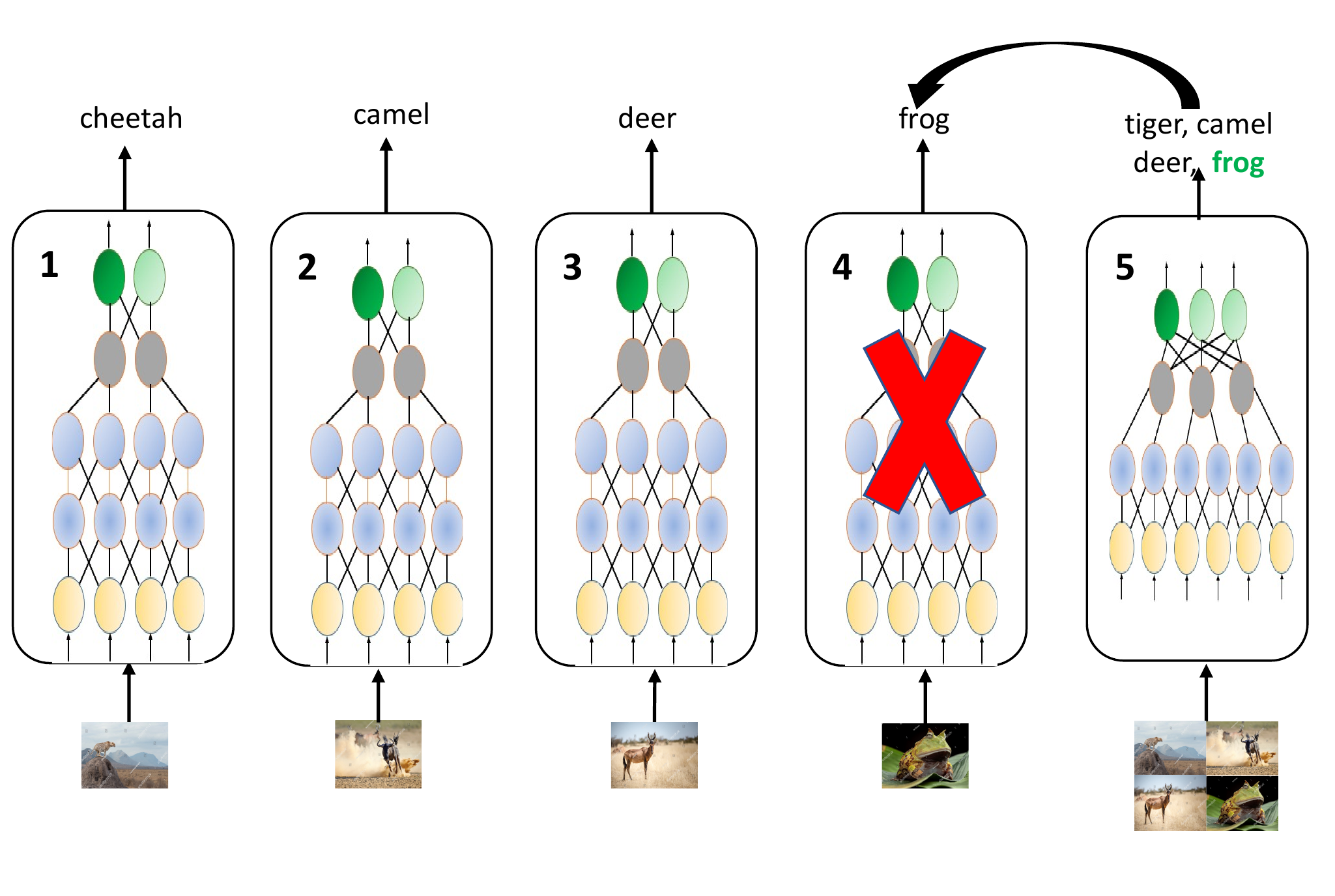}
        \caption{Collage inference}
        \label{fig:Collage}
    \end{subfigure}
    \caption{Introduction}
    \label{fig:introduction_framework}
\end{figure}

We propose the collage inference technique as a cost effective redundancy strategy to deal with variance in inference latency. Collage inference uses a unique convolutional neural network (CNN) based coded redundancy model, referred to as a collage-cnn, that can perform multiple predictions in one shot, albeit at some reduction in prediction accuracy. Collage-cnn is like a parity model where the input encoding is the collection of images that are spatially arranged into a collage as depicted in figure \ref{fig:introduction_framework}. Its output is decoded to get the missing predictions of images that are taking too long to complete. This coded redundancy model is run concurrently as a single backup service for a collection of individual image inference models. We refer to individual single image inference model as an s-cnn model. In this paper, we describe the design of the collage-cnn model and we demonstrate the effectiveness of collage inference on cloud deployments.

The main contributions  of this paper are as follows:
\begin{itemize}
    \item We propose collage-cnn model which performs multi-image classification as a low cost redundant solution to mitigate slowdowns in distributed inference systems.
    \item We describe the design and architecture of the collage-cnn models. We describe techniques to generate large training datasets for training the collage-cnn models.
    \item We evaluate the collage-cnn models by deploying them in the cloud and show their effectiveness in mitigating slowdowns without compromising prediction accuracy. We demonstrate that collage-cnn models can reduce 99-th percentile latency by 1.2x to 2x compared to alternate approaches.
\end{itemize}
Rest of the paper is organized as follows. In section \ref{sec:background} we provide background and motivation. In section \ref{sec:collage_inference} we describe the collage inference techniques. We describe architecture of models and implementation in section \ref{sec:implementation}. Section \ref{sec:evaluations} provides experimental evaluations and design space exploration, section \ref{sec:related_work} discusses related works and in section \ref{sec:conclusion} we draw conclusions.



\section{Background and Motivation}
\label{sec:background}
In this section we provide background on image classification and object detection. We then demonstrate the prevalence of long tail latency while performing image classification in the cloud.
\subsection{Background} 
\label{subsec:background}
\textbf{Image classification:}
Image classification is a fundamental task in computer vision. In image classification, the goal is to predict the main object present in a given input image. There are a variety of algorithms, large datasets, and challenges for this task. A widely known challenge is the ImageNet Large Scale Visual Recognition Challenge (ILSVRC). The training dataset consists of 1.2 million images that are distributed across 1000 object categories. Since 2012 \cite{krizhevsky2012imagenet}, the improvements in accuracy of image classification tasks have come from using Convolutional Neural Networks (CNNs). Some of the popular CNN architectures are: ResNet \cite{resnet2016}, Wide ResNet \cite{wide_resnet2016}, Inception \cite{inception2015}, MobileNet \cite{mobilenet2017}, VGGNet \cite{vggnet2014}.


\begin{figure}
    \centering
    \includegraphics[width=\linewidth]{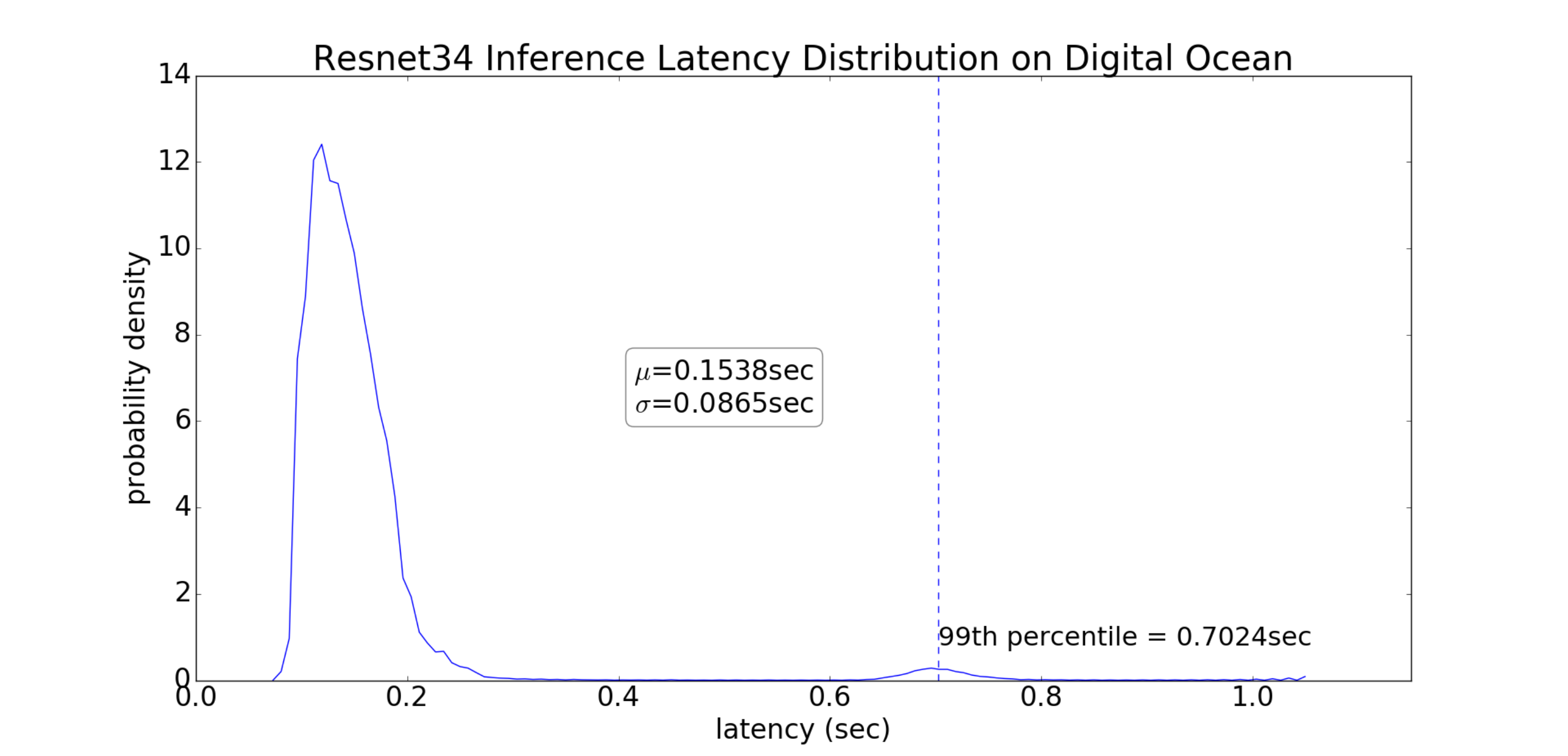}
    \caption{Inference latency distribution}
    \label{fig:motivation_inference_distribution}
\end{figure}

\textbf{Object detection:}
Given an input image, the object detection task involves predicting two things: the classes of all objects present in the image, the locations of objects in the image. The location information is predicted as a rectangular bounding box within the image. 
There are two methods to perform object detection using CNNs: region based detectors predicting object locations in one stage followed by object classification in the next stage \cite{rcnn, fast_rcnn, faster_rcnn, rfcn}, unified or single shot object detection \cite{yolov1, yolov3, ssd2016, dsod2017, dssd2017}. The single shot object detection models have lower inference latency while maintaining similar accuracy as that of the region based detectors.

\begin{figure*}
    \centering
    \includegraphics[width=0.9\linewidth]{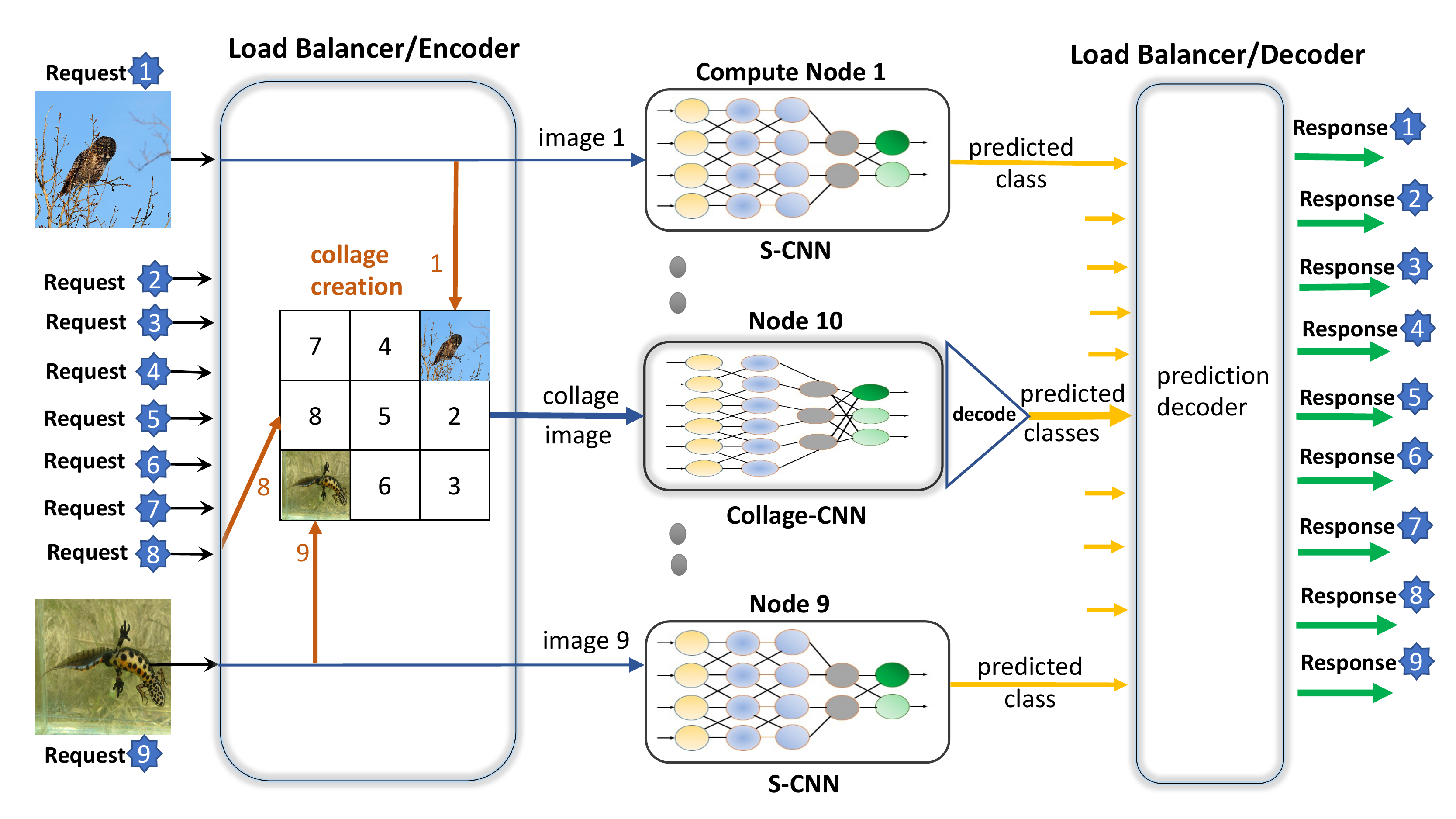}
    \caption{Collage inference algorithm}
    \label{fig:collage_inference_algorithm}
\end{figure*}

\subsection{Characterizing tail latency}
\label{subsec:motivation}
We measured the tail latency for image classification in the cloud. For this purpose, we designed an image classification server that uses a ResNet-34 CNN model to provide inference and created 50 instances of this server on Digital Ocean cloud ~\cite{digital_ocean}. Each instance is running on a compute node that consists of 2 CPUs and 4 GB memory. Clients generates requests to all these servers. We measured inference latency across these 50 nodes while performing single image classification. The probability density function of the latency is shown in figure \ref{fig:motivation_inference_distribution}. The average single image inference latency was 0.15 seconds whereas the 99-th percentile latency was 0.70 seconds. The 99-th percentile latency is significantly (4.67x) higher than the mean latency. 
This long tail in latency degrades Quality of Service (QoS) and impacts Service Level Agreements (SLAs) adversely\cite{tail, information_visualizer}. This motivates the need for techniques that can mitigate slowdowns and lower the variation in latency.  


\section{Collage Inference Technique}
\label{sec:collage_inference}
In this section we describe limitations in existing straggler mitigation techniques and then we describe the collage inference technique.

\subsection{Limitations of existing techniques}
\label{subsec:existing}

One option to improve QoS in the presence of high latency variance or a  straggler node is to add redundancy in the form of over-provisioning of compute nodes. Consider a system of 10 nodes over-provisioned by 1 node. This node would be running another replica of s-cnn. But it is difficult to know ahead of time which one of the $N=10$ nodes will be a straggler. As a result, deciding which one of the input requests to replicate becomes difficult.  One strategy is to duplicate the inference request sent to node $i$ only when the node $i$ is detected as a straggler. This is a reactive approach that requires observing a slowdown before launching a redundant request. For instance,  a request may be launched speculatively after waiting for an expected latency, similar to what is adopted in Hadoop MapReduce frameworks~\cite{mapreduce2008,hadoop}. There are practical challenges in implementing the reactive approach. First, from our measurements, shown in figure \ref{fig:motivation_inference_distribution}, the inference latency could be in 10's to 100 milliseconds. As a result, speculative relaunch techniques must be fast enough to adopt. Second, the image must be re-distributed to a new machine for replicated execution. As a result reactive approach may increase the service latency depending on how long the reactive approach waits for a response before speculating a job. 
To avoid the challenges, the system can be over provisioned by factor of 2. That is for every one of $N$ nodes there will be a backup node and every input request will be duplicated. However, this approach increase the compute and communication costs by 2x.


Another technique using coded computing to address straggler mitigation in distributed inference \cite{learning_a_code} uses learned encoding and decoding neural networks to provide redundancy. Briefly the technique is as follows.\\
In a system consisting of $N=5$ compute nodes 1 node, say $O$, is added for redudancy. Each of the $N$ nodes executes a replica of s-cnn. The model in node $O$ takes as input all the 5 input images. These images are passed through a convolutional encoder network and the outputs are then passed onto the s-cnn model. The outputs from the $N+1$ models are fed to a decoder network, composed of fully-connected layers. The outputs from any straggler node is represented as a vector of zeros. The final output from the decoder network generates the missing prediction. Both the encoder and decoder networks are trained through back-propagation. The training data consists of images and also the expected predictions under different straggler scenarios.

This technique when evaluated on CIFAR-10 dataset shows a recovery accuracy of 80.74\% for $N=2$ nodes, and the recovery accuracy is 64.31\% for $N=5$ nodes, when any one of the $N$ nodes is a straggler. One reason for the significant accuracy loss is that the encoding network does not preserve the spatial information of the individual input images.

\subsection{Collage inference}
\label{subsec:collage_inference_technique}
\textbf{Technique}:
A critical insight behind collage inference is that the spatial information within an input image is critical for CNNs to achieve high accuracy, and this information should be maintained. Hence, we use a collage image composed of all the images as the encoding. The encoding used by collage-cnn is a simple spatial arrangement of images $[Image_1,.., Image_i, .., Image_N]$ in a grid format so as to preserve the individual image information, albeit at a reduced resolution. The collage-cnn model is a novel multi-object classification model. The collage-cnn provides the predictions for all the objects in the collage along with the locations of each object in the collage. The predicted locations are in the form of rectangular bounding boxes. By  encoding the individual images into a collage grid of images and using location information from the collage-cnn predictions, the collage inference technique can replace the missing predictions from any slow running or failed nodes.

Since the goal of our work is to mitigate stragglers using a single collage-cnn model, it is imperative that the collage-cnn which acts as a redundant classification model to be as fast as the single image classification task latency.

\begin{figure*}
    \centering
    \begin{subfigure}{0.24\linewidth}
    \includegraphics[width=\linewidth]{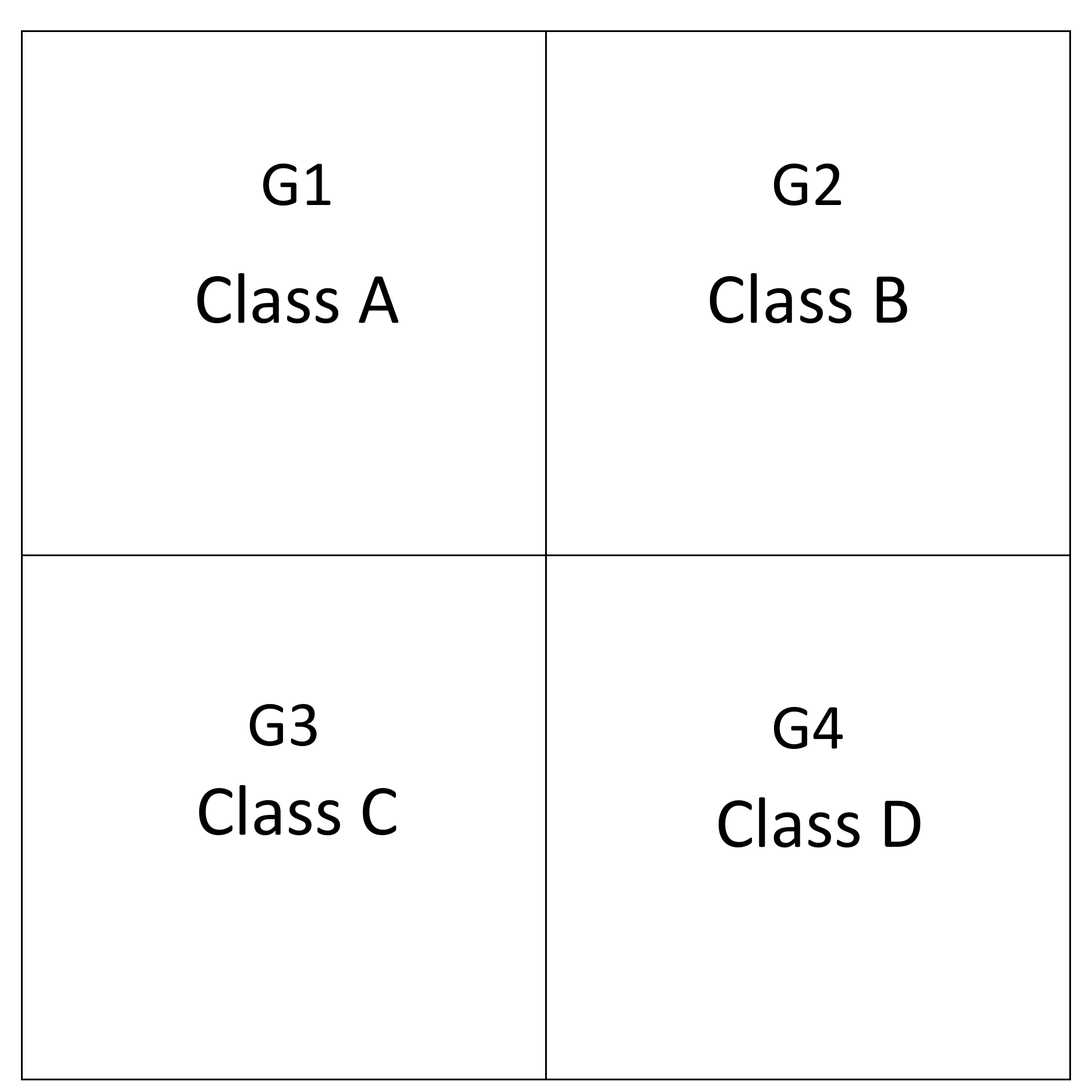}
    \caption{Ground Truth}
    \end{subfigure}
    \begin{subfigure}{0.24\linewidth}
    \includegraphics[width=\linewidth]{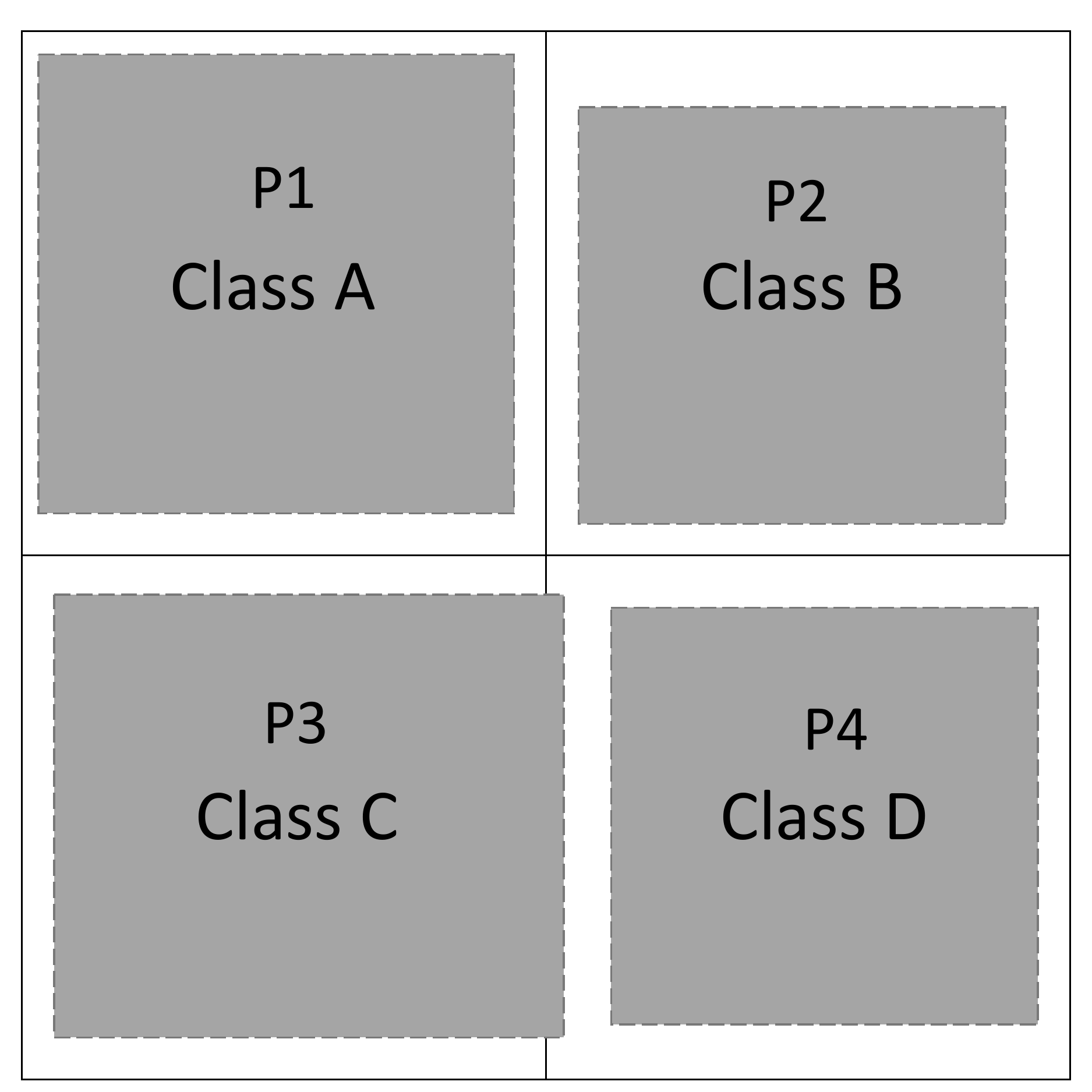}
    \caption{Scenario 1}
    \end{subfigure}
    \begin{subfigure}{0.24\linewidth}
    \includegraphics[width=\linewidth]{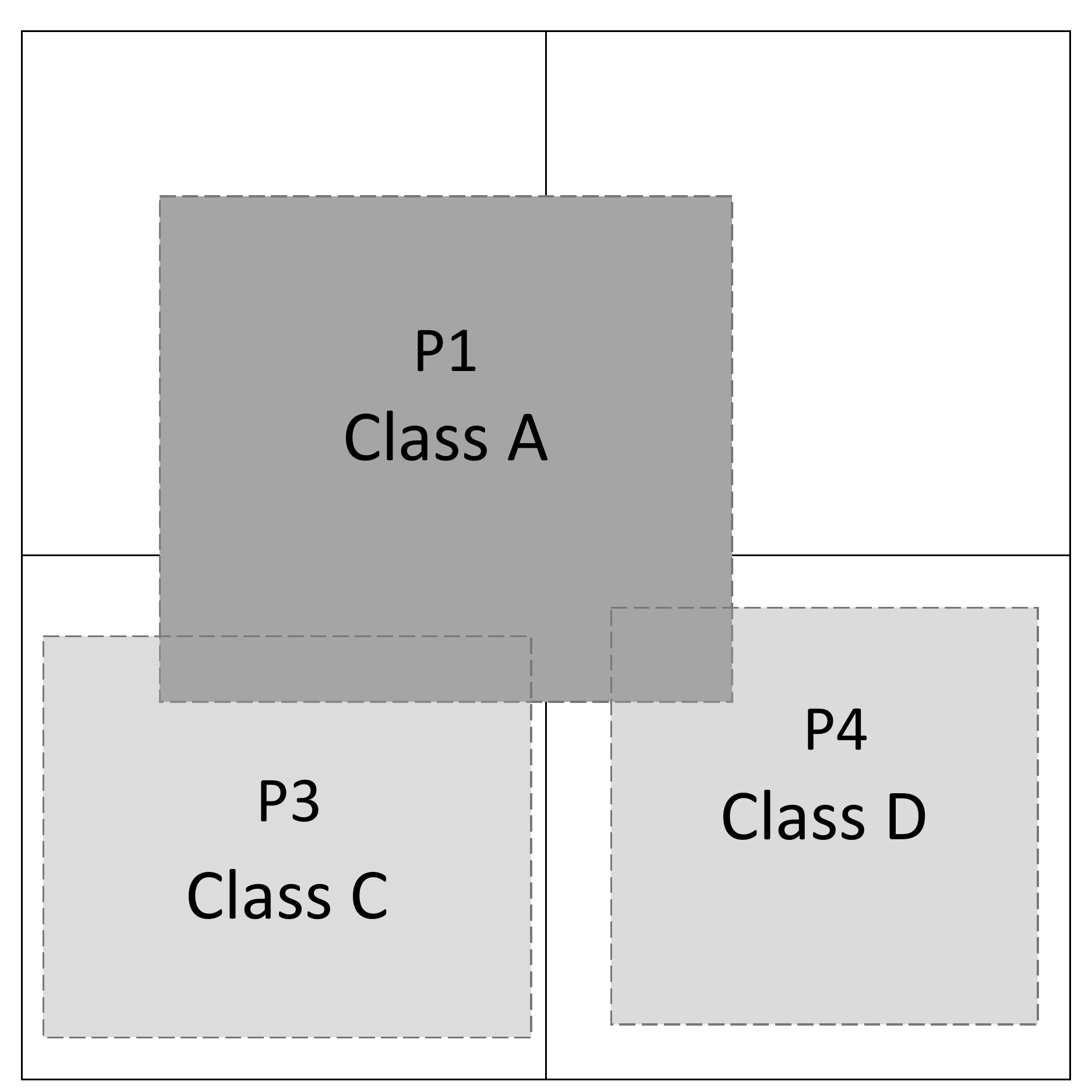}
    \caption{Scenario 2}
    \end{subfigure}
    \begin{subfigure}{0.24\textwidth}
    \includegraphics[width=\textwidth]{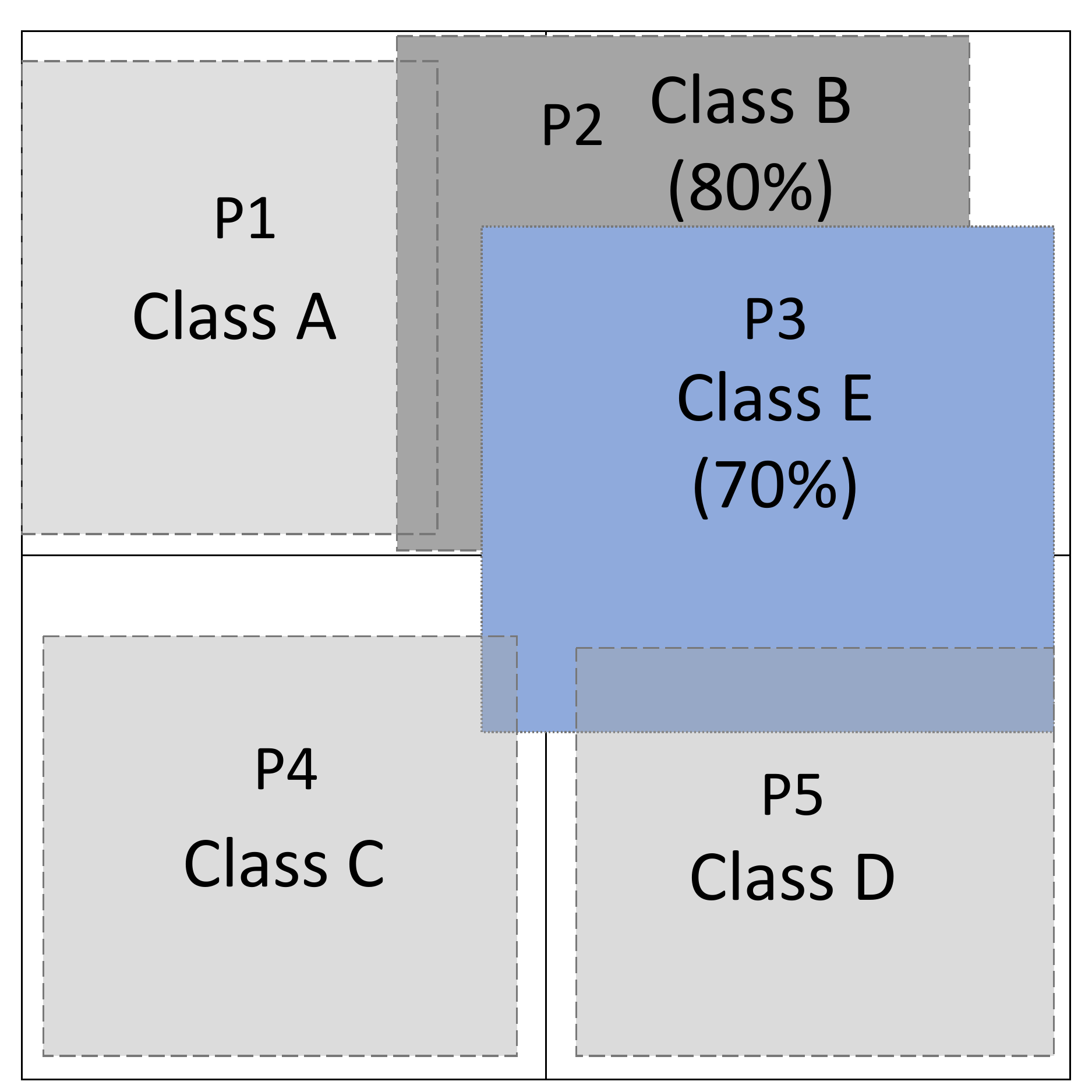}
    \caption{Scenario 3}
    \end{subfigure}
    \caption{Collage decoding scenarios}
    \label{fig:collage_output_scenarios}
\end{figure*}

\textbf{Encoding}:
The encoding of individual images into a single collage image happens as follows. Let a collage-cnn be providing backup for $N$ s-cnn model replicas that are each running on a compute node. To encode the $N$ images into a collage, we first create a square grid consisting of $[\sqrt{N}, \sqrt{N}]$ boxes. Each image that is assigned to an s-cnn model running on compute node $i$ is placed in  a predefined square box within the collage. Specifically, in the collage, each compute node $i$ is assigned the box location $i$. This encoding information is used while decoding outputs of the collage-cnn. From the outputs of the collage-cnn, class prediction corresponding to each bounding box $i$ is extracted using the collage decoding algorithm, and this prediction corresponds to a backup prediction for compute node $i$. As the size of $N$ grows, more images must be packed into the collage, which reduces the resolution of each image and can lower the accuracy of collage-cnn predictions.

\textbf{Example}:
Figure \ref{fig:collage_inference_algorithm} shows a collage inference system consisting of $N=10$ nodes. In this illustration the load balancer receives 9 concurrent requests for image classification. The load balancer manages 10 compute nodes on its back-end. Nine out of the ten nodes run replicas of a s-cnn model for single image classification. The load balancer forwards one inference request to each of these 9 s-cnn models. Concurrently the load balancer also acts as an image encoder by creating a collage from these 9 images. For collage encoding each of the nine input images is lowered in resolution and inserted into a specific location to form the collage image. The input image to node $i$ goes into location $i$ in the collage image. This collage image is provided as input to node 10, which runs the collage-cnn model. The predictions from the collage-cnn are processed using the collage decoding algorithm. The output predictions from the ten nodes go to the load balancer, which processes them and provides the final 9 predictions. 

\textbf{Decoding}:
The collage decoding algorithm extracts the best possible class predictions for the $N$ images from all the collage-CNN predictions. First, all the predictions with confidence values less than detection threshold are ignored by the algorithm. In our experiments, we use a detection threshold of 0.15. The decoding algorithm calculates the Jaccard similarity coefficient, also referred to as Intersection over Union, of each predicted bounding box with each of the $N$ ground truth bounding boxes that are used in creating the collages. Let area of ground truth bounding box be $A_{gt}$, area of predicted bounding box be $A_{pred}$ and area of intersection between both the boxes be $A_i$. Then jaccard similarity coefficient can be computed using the formula: $\frac{A_i}{A_{gt} + A_{pred} - A_i}$. The ground truth bounding box with the largest similarity coefficient is assigned the class label of the predicted bounding box. As a result, the image present in this ground truth bounding box is predicted as having an object belonging to this class label. This is repeated for all the object predictions. 

To illustrate the algorithm, consider example scenarios shown in figure \ref{fig:collage_output_scenarios}. The ground truth input collage is a 2x2 collage that is formed from four images. It has four ground truth bounding boxes G1, G2, G3, and G4 which contain objects belonging to classes A, B, C, and D respectively. Note that this ground truth bounding boxes are created by the load balancer while encoding a collection of images. In scenario 1, the collage model predicts four bounding boxes P1, P2, P3 and P4 with predicted image labels as A, B, C and D, respectively. In this scenario: P1 would have largest similarity value with G1, P2 with G2, P3 with G3 and P4 with G4. So, the decoding algorithm predicts class A in G1, class B in G2, class C in G3, class D in G4. In scenario 2, three bounding boxes are predicted by the model. Predicted box P1 is spread over G1, G2, G3 and G4. The similarity value of P1 with box G1 is: $\frac{1}{3}$, G2 is: $\frac{1}{7}$, G3 is: $\frac{1}{7}$ and G4 is: $\frac{1}{17}$. So, the algorithm predicts class A in G1, empty prediction in G2, class C in G3, class D in G4. In scenario 3, collage model predicts 5 different bounding boxes. Assigning classes A, C, D to boxes G1, G3, G4 respectively is straightforward. But both box P2 and box P3 have highest similarity values with ground truth box G2. Since box P2 has higher confidence (80\%) than box P3 (70\%), collage decoding algorithm predicts G2 as containing class B.

\textbf{Providing final predictions:}
The outputs from collage decoding algorithm along with predictions from all the s-cnn models are provided to the load balancer. The load balancer provides the final predictions as shown in figure \ref{fig:collage_inference_algorithm}. If the predictions from all the s-cnn models are available, the load balancer just provides these predictions as the final predictions and discards the collage-cnn outputs, since there were no slowdowns. In the case where predictions from any of the s-cnn models is not available i.e., there is a slowdown, then the prediction from the collage-cnn corresponding to that s-cnn model is used. It can be observed that the outputs from collage-cnn model can be used to tolerate more than one request slowdown. The predictions from the collage-cnn model can be used in the place of any missing predictions from the s-cnn models. In the rare scenarios where there is a slow s-cnn node and the corresponding prediction from collage-cnn is empty, or the collage-cnn model is also slow, the request to slow s-cnn model is replicated. Prediction from this replicated request is used by the load balancer process.

\textbf{Resource overheads:}
A 2 x 2 collage-cnn works on a 2 x 2 collage composed from 4 single images. It is used in a system where four individual images are sent to four compute nodes, each running a s-cnn model, while the 2 x 2 collage is sent to the node running the 2 x 2 collage-cnn model. In this system, the overhead of running the collage-cnn model is 25\% compute resources. This overhead can be reduced by using a 3 x 3 collage-cnn where one node provides redundancy  for 9 nodes, each running a s-cnn model. This system has approximately 11\% overhead. As more images are combined into a collage, the overhead of using the collage-cnn reduces. If the size of the collage image is fixed, as more single images are packed into a collage the resolution of each image gets reduced. This can reduce the accuracy of the collage-cnn models. We explore this tradeoff between resource overheads, accuracy and latency of different collage-cnns in the evaluation section.

Next we discuss the architecture of the collage-cnn and s-cnn models, generation of training images and training of collage-cnn.

\section{Collage-cnn Architecture and Implementation}
\label{sec:implementation}
\begin{figure*}
    \centering
    \includegraphics[width=\linewidth]{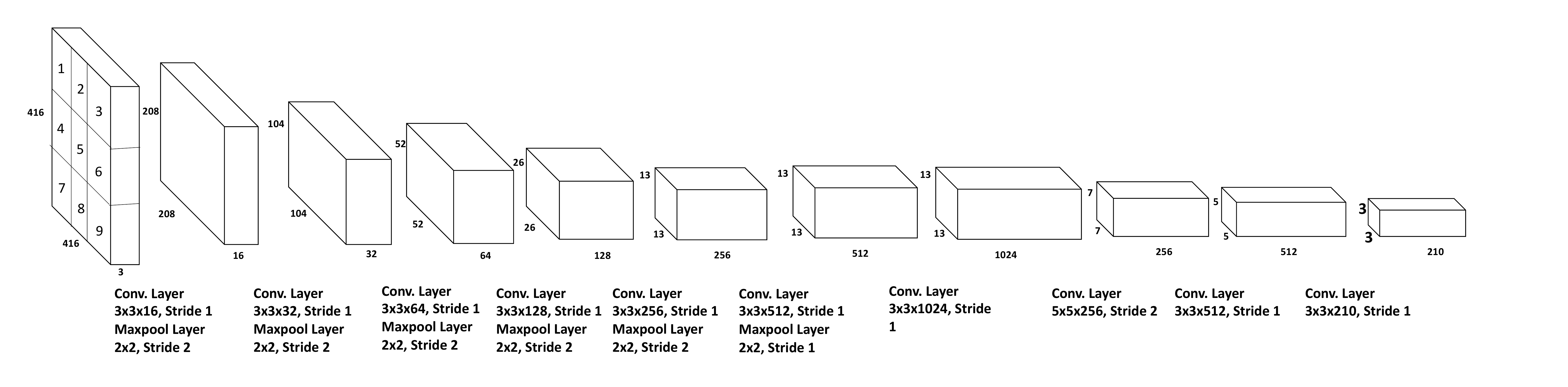}
    \caption{Collage-cnn architecture}
    \label{fig:collage_cnn_network_architecture}
\end{figure*}
\subsection{S-cnn architecture}
We used a pre-trained ResNet-34 model as the the single image s-cnn model for ImageNet-1k dataset. Input to the model is an image of resolution 224 x 224 and the output from the model is one of the 1000 possible class labels. This model has 33 convolutional layers with a fully connected layer at the end to provide class predictions. This model is taken from PyTorch \cite{pytorch} model zoo.
A Pre-trained Resnet-32 model is used as s-cnn model for CIFAR-10 dataset. Input to the model is an image of resolution 32 x 32 and the output from the model is one of the 10 possible class labels. This model has 31 convolutional layers with a fully connected layer at the end to provide class predictions. This model is taken from Tensorflow \cite{tensorflow2016} model zoo.
Both the s-cnn models are out of the box pre-trained models.

\subsection{Collage-cnn architecture}
During inference, collage-cnn acts as the backup classification model for a collection of individual inference models. This requirement places a design constraint on the collage-cnn. The latency of the collage-cnn should be lower than or equal to the latency of s-cnn. Since we use ResNet-34 as the s-cnn, the collage-cnn should have a latency lower than ResNet-34. Our collage-cnn architecture is inspired by the yolov3-tiny model architecture \cite{yolov3}, which is a fast single shot object detection model. Collage-cnn architecture consists of a total of 10 convolutional layers. The final 3 of the 10 convolutional layers are adapted depending on the grid size of the input collage image. The input resolution to our collage-cnn model is 416 x 416 pixels. The final output of the network is a K x K x 210 tensor of predictions. The value of K is the grid dimension of the input collage. If the shape of the input collages is 3 x 3, then the output of the network is 3 x 3 x 210. Each output grid cell predicts two bounding boxes and confidence scores for each of these boxes. Each bounding box prediction is of the form [x, y, width of the box, height of the box, object confidence score, conditional class probabilities]. The (x,y) coordinates correspond to the center of the bounding box. For 100 classes (the total number of image classes we used in this study), the total number of predictions per grid cell is 210. The full network architecture for the 3x3 collage-cnn model is shown in figure \ref{fig:collage_cnn_network_architecture}. Unlike yolov3-tiny model, there are no residual connections and upsampling layers in collage-cnn. Yolov3-tiny uses these layers to detect extremely small objects that may be present in an input frame. But in collage-cnn the objects to be classified in collage images are large enough and there is no need for fine-grained object detection. The collage-cnn is trained using a loss function based on the yolo \cite{yolov1} loss function. The collage-cnn loss function penalizes errors in object confidence, classification and bounding box predictions.

The collage-cnn outperforms yolov3-tiny model while classifying collage images. Customization of the network architecture enables the collage-cnn to have higher top-1 accuracy on collage images. The 3x3 collage-cnn, described in figure \ref{fig:collage_cnn_network_architecture} is 1.4\% more accurate than the yolov3-tiny model while classifying 3x3 collage images. The 4x4 collage-cnn and 5x5 collage-cnn models are also 1.4\% more accurate while classifying 4x4 and 5x5 collage images respectively. Since the collage-cnn has lower number of layers than the yolov3-tiny model, the inference latency of the collage-cnn is 20\% lower.


\subsection{Training of collage-cnn models}
\label{subsec:training_data}
The datasets we used in our experiments are CIFAR-10 and ImageNet-1k (ILSVRC-2012). CIFAR-10 dataset consists of 60000 images divided into 10 object classes. 50000 images are provided for training and 10000 images for validation. ImageNet-1k dataset consists of 1.2 million images divided across 1000 object classes for training and 50000 images for model validation. To train the collage-cnn, collages generated from images of the training datasets are used. To validate the collage-cnn, collages generated from images in validation datasets are used. In our experiments with ImageNet-1k dataset, we use all the training and validation images belonging to 100 of the 1000 classes for evaluations. The selected 100 classes correspond to 100 different objects. We use 100 classes so that collage-cnn model can be trained in a reasonable time, and to explore the design space using limited compute resources.

For the CIFAR-10 based collage dataset we uniformly and at random pick $N$ images from the 50000 training images to create each collage in the training dataset. 
For the ImageNet-1k based collage dataset we first pick all the training images from the 100 classes. Then, we uniformly and at random pick $N$ classes from the 100 classes. One image from each of these $N$ classes is picked and all the $N$ images are combined into a single image. This image is resized to generate the collage image. The $N$ classes need not all be different and some collages have multiple images belonging to same class.

The total possible number of collage images that can be generated is much larger than the number of training images in the raw datasets. This is because there are many permutations to choose from while combining different images into collages. This leads to two advantages. First, it increases the size of the collage-cnn training dataset. Since the task being learned by the collage-cnn is more challenging than the single image s-cnn models, a larger training dataset can help increase the model's accuracy. Second, by generating more number of collage images for training, we try to prevent the model from learning any spurious and fictitious inter-object correlations. In collage-cnn based classification, objects belonging to any class can be present in any location in the image, unlike in object detection.


In our experiments, the input resolution to collage-cnn  model is set to 416 x 416 pixels. So, while forming collages each single image resolution is set to $\frac{416}{\sqrt{N}}, \frac{416}{\sqrt{N}}$ pixels. For the CIFAR-10 dataset, since each image is of the resolution of 32 x 32 pixels, the single image resolution is not reduced even for large $N$ values. For ImageNet-1k dataset, the resolution of single images is lowered even in the 2 x 2 collages, since each image is of the resolution of 224 x 224 pixels. We use the python imaging library to lower the resolution of each image before forming the collage.

For each collage image, the target output of the collage-cnn model consists of $N*5$ values. For each of the $N$ images in the collage there are 5 target outputs: class label, x-coordinate of center of the bounding box, y-coordinate of center of bounding box, width of bounding box, height of bounding box. Given a raw dataset of training images, a python script generates the collage images by appropriately combining single images and scaling down the collage image size. The script also generates the 5 target values for each image in the collage.

\section{Experimental Evaluation}
\label{sec:evaluations}
In this section we first present the accuracy of collage-cnn compared to the ResNet models. We then discuss the end-to-end system performance using collage-cnn. We end by providing comparison of collage-cnn model to alternative redundancy models.

\subsection{\textbf{Training Parameters}}
The models are trained for 130K iterations using Stochastic Gradient Descent (SGD) with the following hyper parameters: learning rate of 0.001, momentum of 0.9, decay of 0.0005, and batch size of 64. While training collage-cnn on ImageNet collages of shapes 4x4 and 5x5 a learning rate of 0.005 is used since using 0.001 caused divergence in SGD. Each model training is performed on a single compute node consisting of a GeForce Titan 1080 GPU equipped with 11 GB of GDDR, 32GB of DRAM and an AMD Ryzen 3 1200 quad-core Processor. The training run time is \textasciitilde26 hours for 130K iterations.

\subsection{\textbf{Accuracy of collage-cnn}}
\textbf{Effects of increasing the training data:} Size of the collage-cnn training data can be increased using the different permutations possible when generating collages, as described in section \ref{sec:implementation}. We performed experiments to measure the effects of using more collage images during training. We observe consistent improvements in validation accuracy.\\
\textbf{1}: While training a collage-cnn model using 4 x 4 ImageNet collages, as the training set size is doubled from 52K (52000) to 104K images, validation accuracy increased by 6.95\%.\\
\textbf{2}: While training a collage-cnn model using 3 x 3 ImageNet based collages, as the training set size is doubled from 26K to 52K images, the validation accuracy increased by 1\%.\\
\textbf{3}: While training a collage-cnn model using 3 x 3 CIFAR-10 based collages, as the training set size is increased from 10K to 50K images, the validation accuracy increased by 1.38\%.\\
While training the collage-cnn models, the number of images across all collages is larger than the number of single training images present in the corresponding dataset. For instance, while training a collage-cnn with CIFAR-10 dataset we created 50,000 collages. For ImageNet the total number of single training images in the 100 classes is 120K. For training the collage-cnn models 208K collages are used.

\begin{figure}
    \centering
    \includegraphics[width=\linewidth]{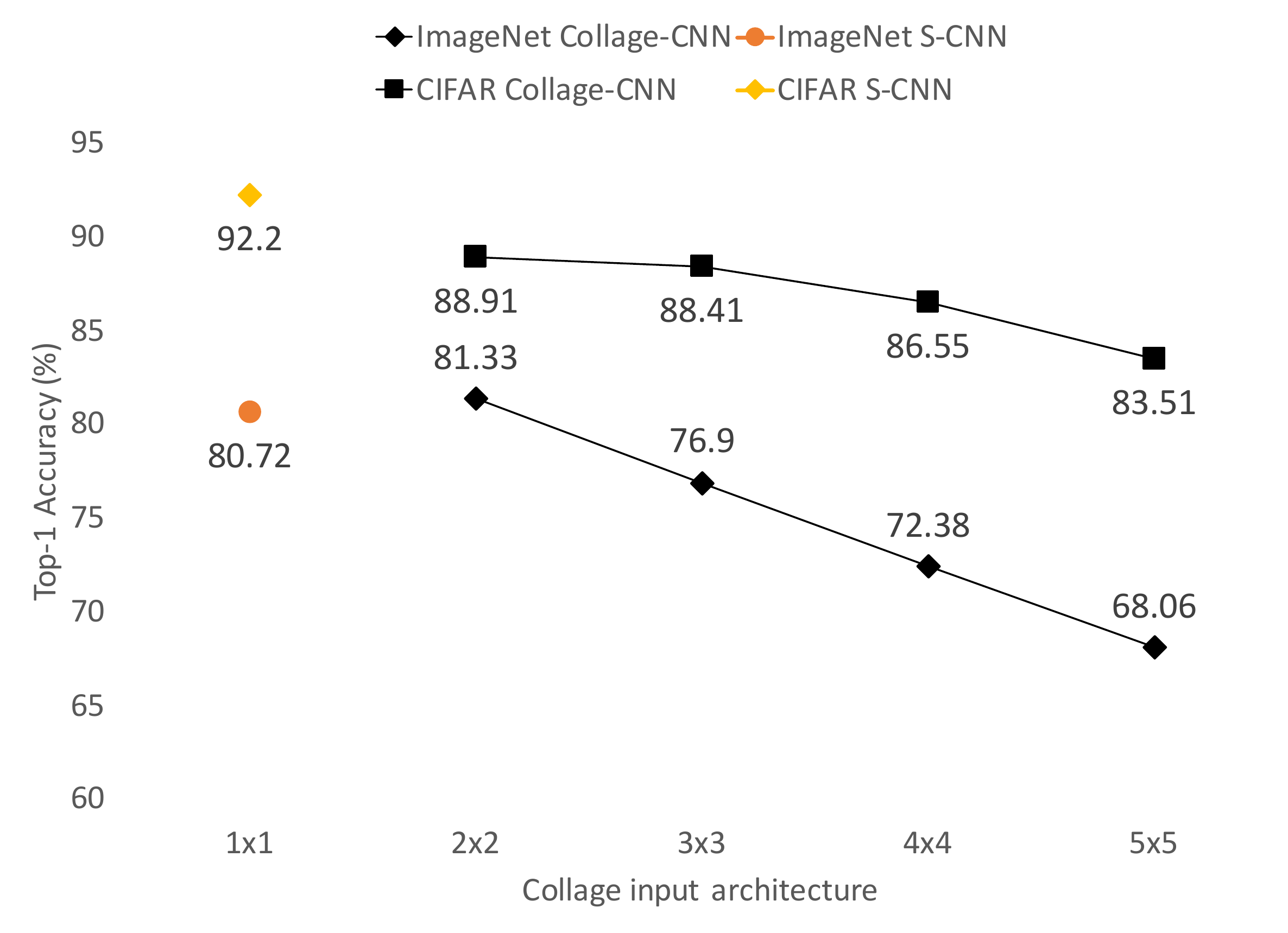}
    \caption{Accuracy on 100 classes of ImageNet-1k}
    \label{fig:accuracy_imagenet}
\end{figure}

\textbf{CIFAR-10 Dataset:} We measured the top-1 accuracy of collage-cnn and s-cnn models using validation images from CIFAR-10. 
The accuracy results are plotted in figure \ref{fig:accuracy_imagenet}. The baseline s-cnn model has a accuracy of 92.2\% whereas the 2x2 collage-cnn models has a accuracy of 88.91\%. Further, it can be seen that the accuracy of collage-cnn models decreases gradually as the number of images per collage increases. 
As stated earlier, when using the CIFAR-10 dataset the collage image resolution was not lowered, since even a 5 x 5 CIFAR images can be fitted in a collage. Hence the gradual loss of accuracy is due to the number of objects that must be detected by the collage-cnn increases and the learning task of collage-cnn model becomes more challenging.

\textbf{Imagenet Dataset:}
Next we measured the top-1 accuracy of collage-cnn and s-cnn models using validation images from ImageNet. Validation collages are generated using the validation images from the 100 ImageNet classes. The resolution of each validation image is lowered to fit into the collage. This is because each validation image has a resolution of 224 x 224 and the collage image resolution is 416 x 416. The top-1 accuracy results are plotted in figure \ref{fig:accuracy_imagenet}.
The accuracy of collage-cnn model for 2 x 2 collages is similar to that of the baseline s-cnn model. This is likely because  resolution of a single image is only slightly reduced while generating 2 x 2 collages. As the number of images per collage increases further the accuracy of collage-cnn model decreases gradually. It can be observed that the rate of decrease in accuracy of collage-cnn model on ImageNet is higher compared to CIFAR-10. As the number of single images per ImageNet collage is increased, resolution of each image gets reduced significantly unlike with CIFAR-10. Hence, the reduced image resolution compounded the complexity of detecting more objects.

\begin{figure*}
    \centering
    \begin{subfigure}{\linewidth}
    \includegraphics[width=\linewidth]{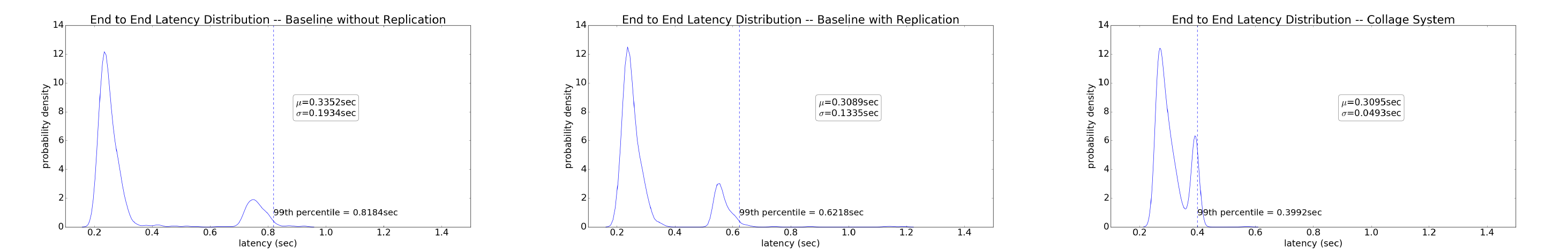}
    \caption{Comparison using 9 s-cnn models in two baselines with a 3x3 collage-cnn (9 s-cnn + 1 Collage-cnn)}
    \end{subfigure}
    \begin{subfigure}{\linewidth}
    \includegraphics[width=\linewidth]{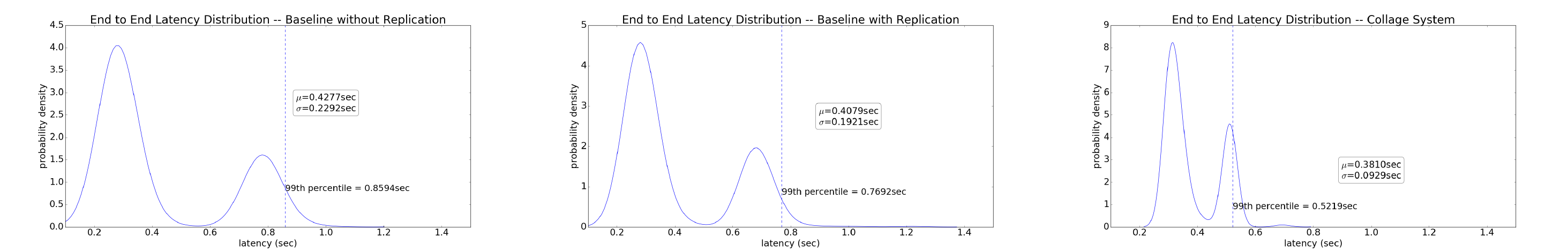}
    \caption{Comparison using 16 s-cnn models in baseline with a 4x4 collage-cnn (16 s-cnn + 1 Collage-cnn)}
    \end{subfigure}
    \begin{subfigure}{\linewidth}
    \includegraphics[width=\linewidth]{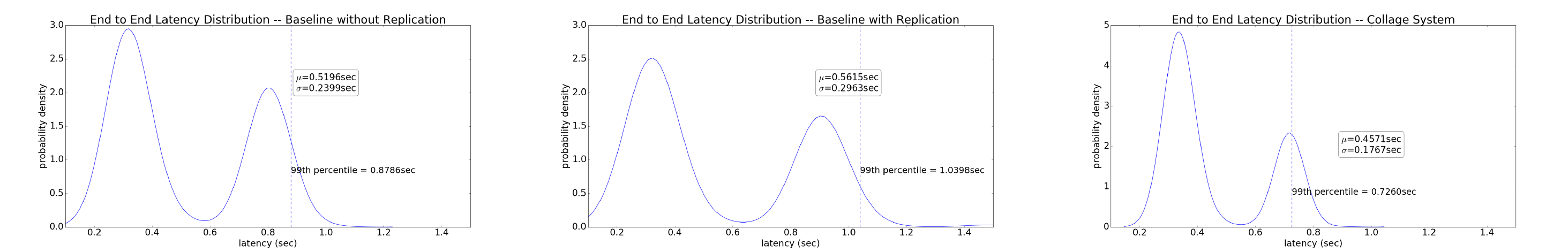}
    \caption{Comparison using 25 s-cnn models in baseline with a 5x5 collage-cnn (25 s-cnn + 1 Collage-cnn)}
    \end{subfigure}
    \caption{Latency distribution from the cloud experiments}
    \label{fig:cloud_experiments}
\end{figure*}

\subsection{Stand alone inference using collage-cnn}
On a cloud server with 2 CPUs and 4 GB of memory, we measure latency of the collage-cnn model. It has \textasciitilde10\% lower inference latency than the s-cnn model. The inference latency is \textasciitilde 0.14 seconds. The reason for collage-cnn inference being faster than the s-cnn inference is that the collage-cnn uses a wider and shallower network; wider network enables more parallelism and shallow layers reduce serial dependencies.

The latencies for encoding images into collages for 3 x 3, 4 x 4 and 5 x 5 collage-cnn models are 0.01, 0.013, and 0.017 seconds respectively. Corresponding collage decoding times are 0.01, 0.028, and 0.047 seconds respectively. Both encoding and decoding times increase as the number of images per collage increases. However, they are significantly smaller than the inference latency.

\subsection{End-to-end system performance with collage-cnn}

We implemented an online image classification system and deployed it on the Digital Ocean cloud \cite{digital_ocean}. The system consists of a load balancer front node, multiple server nodes running s-cnn and collage-cnn models. The load balancer front node performs multiple tasks. It collects requests from clients and generates single image classification requests to the s-cnn models. It also creates a collage from these single images and sends collage classification request to the collage-cnn. It can replicate any single image requests if necessary. We use one Virtual Machine (VM) to host the front node and $N$ additional VMs to serve requests using the s-cnn and collage-cnn models. 
We performed experiments with $N=9, 16, 25$ nodes running s-cnn models and 1 node running collage-cnn. Inference requests are generated using the validation images from the ImageNet dataset.

We compare collage inference with two baseline methods. Implementation of each baseline is as follows.\\
\textbf{1}: First method is where the front node sends requests to the s-cnn servers and waits till all of them respond. The front node does not replicate any slow and pending requests. This is the no replication method.\\
\textbf{2}: In the second method, the front node sends requests to the s-cnn servers with a fixed timeout on all requests. If a server is experiencing a slowdown and does not provide prediction before the timeout, the request is replicated. This is the replication method.\\


Figure \ref{fig:cloud_experiments} shows the end to end latency distribution of the three methods under different levels of redundancy. For requests to the collage-cnn model, the end-to-end latency includes time spent in forming the collage image. The blue curve lines show the estimated probability density function of the end to end latency calculated using Kernel Density Estimation.

\textbf{9 s-cnn + 1 collage-cnn}: The collage inference system has 11\% replication overhead. The mean latency of the collage inference is similar to the replication method and lower than the no replication method. The standard deviation of latency of collage inference is 3.9x and 2.7x lower than the no replication and replication methods respectively. The 99-th percentile latency is 2x and 1.6x lower than the no replication and replication methods respectively.

\textbf{16 s-cnn+ 1 collage-cnn}: The replication overhead of this collage system is 6\%. The mean latency of the collage system is lower than both replication and no replication methods. The standard deviation of the inference latency is 2.5x and 2.1x lower than the no replication and replication methods respectively. The 99-th percentile latency of the inference system is 1.6x and 1.5x lower than the no replication and replication methods respectively.

\textbf{25 s-cnn+ 1 collage-cnn}: This collage inference system has 4\% overhead of replication. The mean latency of the inference system is significantly lower than both the baselines. The standard deviation of latency of the system is 1.36x and 1.68x lower than the no replication and replication methods respectively. The 99-th percentile latency of the collage system is 1.2x and 1.4x lower than the no replication and replication methods respectively.

\textbf{Recovery accuracy}: The total requests sent to the 3x3 collage-cnn during the deployment are shown in row 1 of table \ref{tab:precision_3x3}. For each of that request to the collage-cnn, 9 single image classification requests are sent to the s-cnn models. Thus a total of 74,790 s-cnn requests were measured. When predictions from the collage-cnn are used by the load balancer in place of any unavailable s-cnn predictions, the accuracy of the replacement predictions is 87.4\%. This is referred to as the recovery accuracy and is different from the top-1 accuracy of 78.3\%. The difference comes from the fact that top1-accuracy is the accuracy across all the collage-cnn predictions whereas the recovery accuracy is the accuracy of a subset of predictions used in the place of unavailable s-cnn queries during the deployment. We expect that over the course of many deployments, recovery accuracy would converge to top-1 accuracy. The usage of predictions from the 4x4 collage-cnn model  and their accuracy are shown in table \ref{tab:precision_4x4}.

\begin{table}
\centering
\caption{Usage of the 3x3 collage-cnn model}
\begin{tabular}{|c|c|}
\hline
Total requests & 8310\\
\hline
Total requests that encountered slowdowns & 1480\\
\hline
Collage-cnn predictions are unavailable & 200\\
\hline
Collage-cnn predictions are available & 1280\\
\hline
Collage-cnn predictions are accurate & 1119\\
\hline
Accuracy of used collage-cnn predictions & 87.4\%\\
\hline
\end{tabular}
\label{tab:precision_3x3}
\end{table}

\begin{table}
\centering
\caption{Usage of the 4x4 collage-cnn model}
\begin{tabular}{|c|c|}
\hline
Total requests & 4680\\
\hline
Total requests that encountered stragglers & 1002\\
\hline
Collage-cnn is one of the stragglers & 313\\
\hline
Collage-cnn is not a straggler & 689\\
\hline
Collage-cnn predicted accurately &563\\
\hline
Accuracy of used collage-cnn predictions & 81.71\%\\
\hline
\end{tabular}
\label{tab:precision_4x4}
\end{table}

\subsection{Comparison to alternate backup models}
Apart from the above evaluated two baselines, we further compared 3 x 3 collage-cnn model with  alternative redundancy models. These experiments use ImageNet dataset.

\textit{Multi-image batch ResNet}: In this study we use the ResNet-34 model as the redundancy model but with batching capability of 9 images per single batch. Inference latency of this model is 6.1x larger than using a 3x3 collage-cnn. Batch processing is known to reduce the delay by effectively exploiting any underutilized compute resources.

\textit{Lower resolution CNN}: In this study the CNN used has an architecture that is similar to a 3 x 3 collage-cnn; same number of convolution and pooling layers with the same filter configuration but takes an image with resolution of 139 x 139 as input. 139 x 139 is same as the final resolution of each image in a 3 x 3 collage. Batch inference with nine 139 x 139 single images takes 1.8x longer than the 3 x 3 collage-cnn.

\textit{Multi-image batch MobileNet-v2}: In this study we use multi-image batch Mobilenet-v2 as the redundancy model instead of 3 x 3 collage-cnn. At the full image resolution of 224 x 224 it's top-1 accuracy of 81\% is similar to ResNet-34. However at the lower resolution of 139 x 139, which is the resolution of each image in a 3 x 3 collage, the accuracy shows a steep decline to 71.4\%. A 3x3 collage-cnn on the other hand has a much higher top-1 accuracy of 78.3\%. The latency of doing inference, with 9 lower resolution images per batch, using MobileNet-v2 is 2.6x larger than the latency of doing inference on one 3 x 3 collage image with a collage-cnn. Note that if the resolution is not lowered the latency would be closer to the Resnet-34 batch inference latency. 

\textit{Knowledge distillation}: We trained compressed models with knowledge distillation using s-cnn (ResNet-34) as the teacher model. With a student model that is similar to a 3x3 collage-cnn, classification accuracy is 70.5\%. If the student model has 5 convolutional and 1 fully connected layers, it's accuracy is even lower at 57\%. Accuracy of 3x3 collage-cnn is higher at 78.3\%.

The experimental results presented in this section demonstrate the effectiveness of collage inference over alternate redundancy methods.


\section{Related Work}
\label{sec:related_work}

\noindent \textbf{Tail latency in distributed systems}: Paragon \cite{paragon} presents a QOS aware online heterogenous datacenter scheduler. Techniques proposed in \cite{quasar, dLo, jLeverich, compactorHarchol} focus on improving resource efficiency while providing low tail latency. These techniques are orthogonal to  collage inference technique. Using replicated tasks to improve the response times has been explored in \cite{clones,nihar,wang,gardner,chaubey,lee}. This approach needs multiple replicas of all the data and adds large compute and storage overheads. Another strategy used for straggler mitigation is arriving at an approximate result without waiting on the stragglers \cite{approx}. Ignoring the stragglers is not well suited for distributed inference since it causes significant reduction in accuracy.

\noindent \textbf{Straggler mitigation in distributed  training}: Parameters servers are used in distributed SGD based training. To mitigate failures, distributed parameter server approaches use chain replication of the parameter servers \cite{MuLiOSDI2014}\cite{MuLiNIPS2014}. In this replication when there is  failure of a parameter server A, its load is shifted onto another parameter server B. This approach is similar to the reactive replication approach of Hadoop and has the same drawbacks. Parameter server B may now become a bottleneck to training due to increased communication to and from it.

\noindent \textbf{Coded computation}: Coded computation methods have been proposed to provide resiliency to stragglers in distributed machine learning training ~\cite{speedUpML,reisizadehmobarakeh2017coded,LMA16_unify,dutta2016short,polyCodes, s2c2}. All these methods target linear machine learning algorithms and not non-linear computations like deep neural networks. A concurrent work \cite{parity_models} proposes a general parity models framework, ParM, for coding-based resilient inference in tasks such as image classification, speech recognition and object localization. Similar to collage-cnn models, ParM proposes using parity models as backup models to reconstruct unavailable predictions from slow nodes. The framework allows for the parity model to be different for different inference tasks. During evaluations for image classification, ParM uses parity models having the same architecture as the models they are backing up. In contrast, collage-cnn models use a custom architecture optimized for multi object classification. Using the custom architecture leads to significant increase in the classification accuracy. A 2 x 2 collage-cnn working on 4 CIFAR-10 images has a classification accuracy of 88.91\% whereas the corresponding parity model in ParM (k = 4) has an accuracy of 74\%. Due to this collage-cnn models can provide a much better accuracy using the same compute resources as ParM, or provide a similar accuracy as ParM using much lower compute resources. 

\section{Conclusion}
\label{sec:conclusion}
Cloud based prediction serving systems are being increasingly used to serve image classification based requests. Serving requests at low latency, high accuracy and low resource cost becomes very important. In this paper we described collage inference where a coded redundancy model is used to reduce the tail latency during inference while maintaining high accuracy. Collage inference uses novel collage-cnn models to provide recovery from slowdown during runtime. Collage-cnn models provide good tradeoff between accuracy, resource cost and tail latency. Deploying the models in the cloud we demonstrate that the 99-th percentile latency can be reduced by upto 2x compared to replication based approaches while maintaining high prediction accuracy. We conclude that collage inference is a  promising approach to mitigate stragglers in distributed inference. Our future work includes extending the coded redundancy approach to more deep learning applications.



\bibliography{collage_inference.bib}
\bibliographystyle{sysml2019}

\end{document}